# Leveraging Large Language Models to Predict Antibody Biological Activity Against Influenza A Hemagglutinin

Ella Barkan[1], Ibrahim Siddiqui[2], Kevin J. Cheng[3], Alex Golts[1], Yoel Shoshan[1], Jeffrey K. Weber[3], Yailin Campos Mota[4], Michal Ozery-Flato[1*†], Giuseppe A. Sautto[4*†]

[1] IBM Research-Israel, Haifa, Israel.

[2] Case Western Reserve University, Cleveland, OH, USA.

[3] IBM TJ Watson Research Center, Yorktown Heights, NY, USA.

[4] Florida Research and Innovation Center, Cleveland Clinic, Port St. Lucie, FL, USA.

[*] These authors jointly supervised this work

[†] Co-corresponding authors: Michal Ozery-Flato (ozery@il.ibm.com) and Giuseppe A. Sautto (sauttog@ccf.org).

## Abstract

Monoclonal antibodies (mAbs) represent one of the most prevalent FDA-approved modalities for treating autoimmune diseases, infectious diseases, and cancers. However, discovery and development of therapeutic antibodies remains a time-consuming and expensive process. Recent advancements in machine learning (ML) and artificial intelligence (AI) have shown significant promise in revolutionizing antibody discovery and optimization. In particular, models that predict antibody biological activity enable in-silico evaluation of binding and functional properties; such models can prioritize antibodies with the highest likelihoods of success in costly and time-intensive laboratory testing procedures. We here explore an AI model for predicting the binding and receptor blocking activity of antibodies against influenza A hemagglutinin (HA) antigens. Our present model is developed with the MAMMAL framework for biologics discovery to predict antibody-antigen interactions using only sequence information. To evaluate the model's performance, we tested it under various data split conditions to mimic real-world scenarios.

Our models achieved an AUROC ≥ 0.91 for predicting the activity of existing antibodies against seen HAs and an AUROC of 0.9 for unseen HAs. For novel antibody activity prediction, the AUROC was 0.73, which further declined to 0.63–0.66 under stringent constraints on similarity to existing antibodies. These results demonstrate the potential of AI foundation models to transform antibody design by reducing dependence on extensive laboratory testing and enabling more efficient prioritization of antibody candidates. Moreover, our findings emphasize the critical importance of diverse and comprehensive antibody datasets to improve the generalization of prediction models, particularly for novel antibody development.

# Introduction

The influenza A virus is responsible for one of the most prevalent infectious diseases worldwide, with associated influenza infections impacting an estimated 20-40 million people annually in the United States alone [1,2]. Despite its wide prevalence, the influenza A virus remains a significant global health concern due to its rapid mutation rate and potential spillovers from animal reservoirs that enable persistence and continued evolution [3]. In resource-limited countries, seasonal outbreaks of influenza A can result in severe illness and even death [4]. A key factor in the virus's ability to infect host cells is hemagglutinin (HA), the main protein on the viral surface that facilitates binding to host cell receptors. Importantly, HA also serves as a primary target for the immune system: when an antibody binds to HA, that antibody can neutralize the virus and prevent infection [5]. In fact, HA is the primary antigen contained in all the current standard of care for influenza vaccine formulations, serving as the main target for developing immunity to influenza A.

Over the past two decades, we and other groups have characterized a plethora of monoclonal antibodies (mAbs) directed against the influenza HA that are endowed with different breadth of recognition, neutralization, and protection profiles [6–18]. These highly specific antibodies are critical for the development of diagnostic and immunotherapeutic tools for treating diseases spanning cancer, autoimmune, and infectious disease foci. In particular, mAbs have shown great promise in the prevention and treatment of infections caused by respiratory pathogens like the severe acute respiratory syndrome coronavirus 2 (SARS-CoV-2) and the respiratory syncytial virus (RSV) [19].

Though mAbs now constitute almost a third of all newly FDA-approved treatments, therapeutic antibody discovery remains a lengthy and costly process [20]. Researchers' ability to test new antibody formulations *in silico* represents a critical choke-point; predicting antibody binding to influenza A with AI technology offers a groundbreaking opportunity to model immunology and deliver tangible benefits for global health.

In recent years, AI models have opened new pathways for developing therapeutic molecules. AlphaFold architectures [21–23] achieve impressive single-domain protein structure prediction accuracy but encounter significant challenges with predicting antibody–antigen complex structures and, independent of application, require computationally intensive large-scale sampling [23–25]. Biomedical language models, leveraging natural language processing frameworks, have also been successfully applied to modeling the "languages" of proteins [26–28] and antibodies [29–32], enhancing computational methods for antibody design and optimization [32–35]. Rapid identification and characterization of antibodies that bind viral pathogen antigens remains a critical challenge in antibody design.

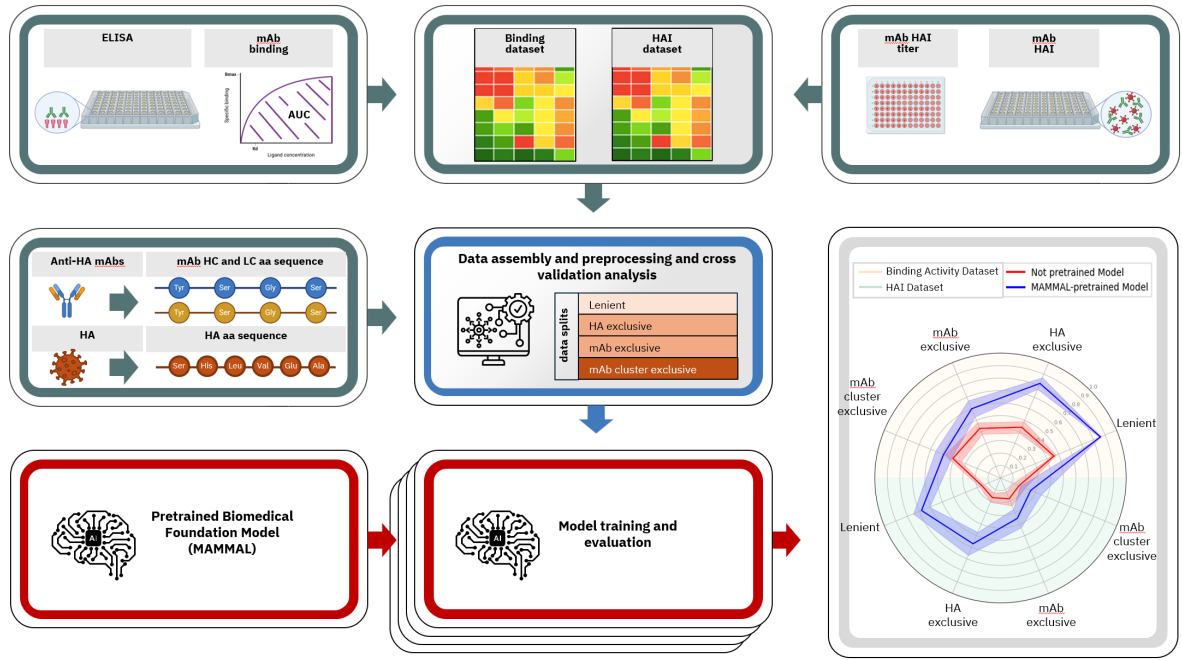

**Figure 1: Study pipeline for developing and evaluating AI models to predict antibody binding and hemagglutination inhibition (HAI) for influenza A hemagglutinin (HA).** The dataset included monoclonal antibody (mAb)-HA amino acid sequences and corresponding ELISA and HAI assay outcomes. To reflect real-world model applications, different data splits were employed, and a cross-validation approach was used to assess model robustness. Predictive models were developed by fine-tuning the MAMMAL biomedical foundation model. Performance was evaluated using the area under the receiver operating characteristic curve (AUROC) and the area under the precision-recall curve (AUPRC), with results compared to those from randomly initialized models. Performance metrics were visualized in a spider diagram for comprehensive comparison.

In this study, we investigate the application of the MAMMAL methodology [19] for developing and fine-tuning pretrained biomedical language models, applying it to predict antibody binding and receptor blocking activity to influenza A hemagglutinin (HA). We fine-tune the pretrained model `ibm/biomed.omics.bl.sm.ma-ted-458m` on two standardized antibody-HA datasets: the first dataset corresponding to HA binding assays, and the second dataset corresponding to hemagglutination inhibition (HAI) assays. Using different data splits, we address three key applications: (1) imputing missing labels for pairs of known antibodies and HAs, (2) predicting activities of known antibodies against new HA sequences, and (3) predicting activities of new and arbitrary antibody sequences against HA sequences. An overview of our study pipeline, from data generation and preprocessing to model development and evaluation, is illustrated in Figure 1.

Our models demonstrated high performance when imputing missing labels or predicting antibody activity against new HA sequences, achieving AUROC scores of ≥0.9 in both scenarios. When predicting anti-HA activity for novel antibodies, our models showed moderate classification power (AUROC of 0.73) that dropped to 0.63–0.66 under stringent antibody similarity constraints. We validated the robustness of our results through cross-validation and extensive subgroup analyses. Furthermore, we assessed the models' abilities to identify broadly protective antibodies.

Our findings highlight the potential of leveraging language models pretrained on large-scale protein and antibody sequence data to develop sequence-based AI models for predicting antibody-target activity, even with a limited labeled dataset. This approach could streamline antibody research by reducing the need for extensive laboratory testing and helping prioritize more expensive *in-silico* screening methods.

## Related Work

Predicting the binding affinity or functional activity of an antibody candidate against a given target antigen represents a key step in therapeutic and vaccine design. Simulation methods like molecular dynamics (MD) and free energy perturbation (FEP) [36–38] can accurately capture the physical states of antibody-antigen complexes and offer one avenue toward making affinity predictions. However, these atomistic simulation approaches are computationally expensive and typically rely on accurate static structural information as a starting point. Molecular docking methods offer higher-throughput but less accurate estimates of intermolecular binding modes; docking scores exhibit notoriously weak correlations with affinity data [39,40]. Other knowledge-based scoring functions [41,42] offer alternative means of estimating binding affinities, but with limited accuracy and validity .

Machine learning (ML) methods (like those based on random forests) have set the standard for affinity prediction for small molecule-target complexes for decades. Similar approaches have since gained prevalence in the antibody discovery space. With the growing availability of large scale molecular datasets, deep learning (DL) techniques have emerged as favored methods for both small molecule-protein [43–45] and antibody-antigen binding prediction [46,47].  The recent trend of using very large protein datasets to pretrain biomolecular LLMs [26,27,48–50], with some models even being specific to antibodies [29–31,51], has redefined antibody-antigen binding affinity prediction as a downstream task. Most approaches in this domain rely on making predictions based on pretrained LLM embeddings or fine-tuning LLMs on labeled data [52].

Experimental three-dimensional structures of antibody-antigen complexes are scarce, expensive, and difficult to solve relative to simple protein sequencing. Breakthroughs in deep learning for protein structural modeling, particularly AlphaFold [21–23] and RoseTTAFold [53], have transformed efforts toward protein structure prediction. Despite significant advances in state-of-the-art structure prediction methods for antibody-antigen complex modeling, further refinements are needed to enhance antibody design efforts [23,54].

In the context of predicting antibody-HA interactions, previous work has relied on various computational approaches. Several studies have employed AlphaFold [55,56] for structural predictions of antibody-HA complexes, aiming to integrate empirical structural biology insights into binding predictions [56]. Other studies [34] have demonstrated the value of protein language model scores in predicting enhanced antibody binding, focusing primarily on antibody sequence analysis without explicit consideration of the target protein.

Several studies have utilized language models for predicting properties of antibodies targeting HA. For example, Wang et al. [57] leveraged a memory B-cell language model to predict novel antibodies targeting the conserved HA stem region. Such applications underscore the potential of fine-tuning of pretrained models for improving antibody-antigen binding discrimination, as demonstrated for both SARS-CoV-2 and influenza antigens [58]. However, most language models lack the ability to directly quantify binding affinity, a task that is critical for accurately predicting and optimizing antibody efficacy.

Prior efforts in predicting antibody HAI titers have largely relied on machine learning models trained on virus-antiserum pairs contained within HAI titer datasets from the Worldwide Influenza Centre (WIC) [59]. Predictive strategies based on Adaboost [60], random forest [61], and Bayesian approaches [62] have achieved notable success in this area. Additional bioinformatics methods [63,64] have provided insights into antigenic drift. Despite these advances, existing methods predominantly focus on virus-antiserum pairs and are not tailored to predict antibody-HA interactions based on sequence data alone.

# Methods

## Monoclonal antibodies, recombinant HA and influenza viruses

The 188 human and mouse anti-HA mAbs used to generate the binding and HAI dataset featured in this study were previously described by our group or derived from BEI and

IRR resources [7–18,65–67]. Recombinant HAs and influenza viruses corresponding to different H1N1 and H3N2 strains were produced as previously described [68,69].

## Data description

This study's dataset comprises pairs of mAbs and influenza A HA antigens, evaluated using two assays: binding activity and HAI activity. These assays characterize 1) the breadth of binding, indicating the number of strains the mAb can bind; and 2) the breadth of receptor binding inhibition (i.e., HAI), which reflects the antibody's ability to prevent viral binding to the sialic acid receptor, thereby blocking virus entry and neutralizing the virus.

Binding activity assays were conducted as previously described [9,11,13,17] using an Enzyme-Linked Immunosorbent Assay (ELISA), a method to detect and quantify antibody-antigen interactions by measuring a signal, such as color change, proportional to binding strength [5]. Binding strength is quantified as the area under the binding curve (AUC-ELISA), derived from 3-fold serial dilutions of mAbs (20 to 0.009 µg/mL). AUC-ELISA values range from 0.5 (negative) to 20, with higher values indicating stronger binding. We considered AUC-ELISA values greater than 1 as positive binding.

HAI represents the ability of an antibody to block the interaction between HA and the sialic acid cell receptor, measured as the minimum antibody concentration required to inhibit red blood cell agglutination by the virus. In particular, the HAI assay involves mixing serially diluted mAbs with a fixed amount of virus and adding red blood cells to detect hemagglutination. Ultimately, the ability of an antibody to inhibit this agglutination is indicative of its neutralizing potency [19]. HAI assays with mAbs were performed as previously described [9–12,17]. HAI values range from 0.005 to 20 µg/mL, with lower values indicating higher potency. We classified HAI values below 10 µg/mL as positive outcomes. The mAbs in the dataset are represented as amino acid sequences of the variable regions of the heavy (HC) and light (LC) chains.

The antigens in the dataset are represented as amino acid sequences of the HA protein. The amino acid sequences for 79% of antigens were obtained from publicly available databases including NCBI GenBank and GISAID. The remaining 21% of antigens were derived from experimental results conducted against previously described COBRA (Computationally Optimized Broadly Reactive Antigen) HA. These include H2_COBRA.Z7[70], H3_NG5 [71], H3_NG7 [72], H5_COBRA2 [73], H1N1 COBRA P1 [74], H1N1 COBRA X3 [75], H1N1 COBRA X6 [76], H3N2 COBRA NG2 [77], H1N1 COBRA Y2 [78], H3N2 COBRA T10 [79], and H3N2 COBRA T11 [80]. COBRA antigens are computationally optimized to elicit broader reactivity than wild-type antigens and were designed to enhance breadth of response by reconciling the variability among different seasonal and/or pandemic HA strains in a single HA antigen [7]. Additional metadata

available for antibodies includes the ISO type of the light and heavy chains, as well as epitope information

## Data splits

We assessed model robustness through 5-fold cross validation across four distinct data splitting strategies: lenient, HA-exclusive, mAb-exclusive, and mAb-cluster exclusive. In the lenient split, mAb-HA pairs were randomly distributed between training and testing sets. For the HA-exclusive split, we ensured all pairs containing the same HA sequence were assigned to the same fold, appearing exclusively in either the training or test set. The mAb-exclusive split followed a similar principle, keeping all pairs involving the same mAb within a single fold. In the mAb-cluster exclusive split, we grouped mAb-HA pairs based on antibody clusters, which were determined using MMseqs2 [81] with a minimum sequence identity threshold of 50%. We generated these four splitting strategies separately for each label type (binding activity and HAI), maintaining consistent proportions of positive antibody-HA pair labels across all folds.

## Binding and HAI classification models

We developed predictive models using MAMMAL [50], a recently published biomolecular foundation model framework. MAMMAL is part of IBM's Biomedical Foundation Modeling (BMFM) technology suite[1]. Specifically, we used the model `ibm/biomed.omics.bl.sm.ma-ted-458m`, which was trained on extensive multi-domain data, including proteins (UniProt [82]), antibodies(OAS [83]), and protein-protein interactions (STRING [84]) via various self-supervised tasks. For brevity, we refer to this foundation model as MAMMAL. The model code and pretrained weights are publicly available at https://github.com/BiomedSciAI/biomed-multi-alignment and https://huggingface.co/ibm/biomed.omics.bl.sm.ma-ted-458m respectively. The developed model take a pair of antibody and antigen sequences as input, as represented by their amino acid sequences and using the prompt syntax of the AbAg Bind task described in past work [50]. Model training and evaluation were conducted using the FuseMedML framework [85].

For each task and training fold, we fine-tuned MAMMAL with the default hyperparameters outlined in [50]: AdamW optimizer with β1 = 0.9 and β2 = 0.999, weight decay of 0.01, and gradient clipping with a norm of 1.0. We employed 2K warm-up steps to reach the maximum learning rate, followed by a cosine decay scheduler that reduces the learning rate to 10% of the maximum by the end of training. The maximum input length was set to 900, ensuring input sequences are not truncates. For all models, training involved 1000 iterations on a V100-32G GPU using batch sizes of 8.

---

[1] https://research.ibm.com/projects/biomedical-foundation-models

Table 1: Binding activity dataset characteristics and their association with positive binding classification.

|  |  | Overall | Positive, N (%) | P-Value |
|---|---|---|---|---|
| **Total** |  | 4922 | 1740 |  |
| **mAb host, n (%)** | Human | 4181 (84.9) | 1357 (78.0) | <0.0001 |
|  | Mouse | 741 (15.1) | 383 (22.0) |  |
| **HA year, n (%)** | 2000-2010 | 1336 (27.1) | 458 (26.3) | 0.02 |
|  | <1950 | 903 (18.3) | 291 (16.7) |  |
|  | >2010 | 2190 (44.5) | 795 (45.7) |  |
|  | other/unknown | 493 (10.0) | 196 (11.3) |  |
| **HA subtype, n (%)** | H1 | 2366 (48.1) | 920 (52.9) | <0.0001 |
|  | H2 | 53 (1.1) | 14 (0.8) |  |
|  | H3 | 2210 (44.9) | 740 (42.5) |  |
|  | H5 | 278 (5.6) | 64 (3.7) |  |
|  | H7 | 15 (0.3) | 2 (0.1) |  |
| **mAb LC ISO, n (%)** | kappa | 2806 (57.0) | 1198 (68.9) | <0.0001 |
|  | lambda | 2116 (43.0) | 542 (31.1) |  |
| **mAb HC ISO, n (%)** | IgA | 490 (10.0) | 213 (12.2) | <0.0001 |
|  | IgG | 4432 (90.0) | 1527 (87.8) |  |
| **mAb Epitope, n (%)** | conformational | 2493 (50.7) | 844 (48.5) | 0.03 |
|  | other/unknown | 2429 (49.3) | 896 (51.5) |  |

## Model Evaluation

We evaluated the models' classification performance using the Area Under the Receiver Operating Characteristic Curve (AUROC) and the Area Under the Precision-Recall Curve (AUPRC). Both metrics have values ranging between 0 and 1 For a random classifier, the expected AUROC is 0.5; the expected AUPRC for a random classifier corresponds to the rate of positivity in the dataset [86]. For each 5-fold cross-validation experiment, we report average performance metrics along with their standard deviations across five folds.

# Results

## Data statistics

The classification dataset consists of 188 mAbs and 79 hemagglutinins HAs. The data include 4,922 unique mAb-HA pairs from 176 mAbs and 59 HAs in the binding activity assays and 5,035 pairs from 186 mAbs and 59 HAs in the HAI assays. Among these, 3,188 mAb-HA pairs are shared between the binding activity and HAI assays, involving

**Table 2: HAI activity dataset characteristics and their association with positive HAI classification.**

|  |  | All, N(%) | Positive HAI, N (%) | P-Value |
|---|---|---|---|---|
| Total |  | 5035 | 572 |  |
| mAb host, n (%) | Human | 4112 (81.7) | 439 (76.7) | 0.002 |
|  | Mouse | 923 (18.3) | 133 (23.3) |  |
| HA year, n (%) | 2000-2010 | 1093 (21.7) | 140 (24.5) | <0.0001 |
|  | <1950 | 1851 (36.8) | 72 (12.6) |  |
|  | >2010 | 1761 (35.0) | 322 (56.3) |  |
|  | other/unknown | 330 (6.6) | 38 (6.6) |  |
| HA subtype, n (%) | H1 | 3048 (60.5) | 233 (40.7) | <0.0001 |
|  | H3 | 1832 (36.4) | 339 (59.3) |  |
|  | H5 | 155 (3.1) |  |  |
| mAb LC ISO, n (%) | kappa | 2955 (58.7) | 263 (46.0) | <0.0001 |
|  | lambda | 2080 (41.3) | 309 (54.0) |  |
| mAb HC ISO, n (%) | IgA | 462 (9.2) | 53 (9.3) | 0.998 |
|  | IgG | 4573 (90.8) | 519 (90.7) |  |
| mAb Epitope, n (%) | conformational | 2517 (50.0) | 257 (44.9) | 0.01 |
|  | other/unknown | 2518 (50.0) | 315 (55.1) |  |

174 mAbs and 39 HAs. Positive pairs account for 35% of the binding dataset and 11% of the HAI dataset. On average, each mAb is involved in 28 ± 17 (mean +/- standard deviation) binding assays, while each antigen is included in 83 ± 45 binding assays. For HAI assays, each mAb is involved in 27 ± 17 assays, and each antigen is included in 85 ± 41 assays.

Tables 1 and 2 summarize the data characteristics and statistical associations with positive outcomes for the binding and HAI datasets, respectively. The majority of assays in the datasets involve antibodies (82–85%) originated from human hosts. H1N1 and H3N2 subtypes dominate, representing 88–95% of the data. The mean lengths of the variable regions in the LC and HC sequences of the mAbs, determined using AbNumber [29], are 122 ± 8 and 108 ± 4, respectively. The mean length of antigen sequences is 559 ± 20.

Clustering the 176 mAbs in the binding dataset and the 186 mAbs in the HAI dataset yielded 76 clusters in total for each dataset, with cluster sizes ranging from 1 to 9. The mean cluster size was 2.2 ± 2.0 in the binding dataset and 2.4 ± 2.1 in the HAI dataset. In the binding activity dataset, the proportion of positive pairs ranged from 28% to 41% across all folds in the 4 splits (20 folds in total). For the HAI dataset, this proportion ranged from 7% to 17%.

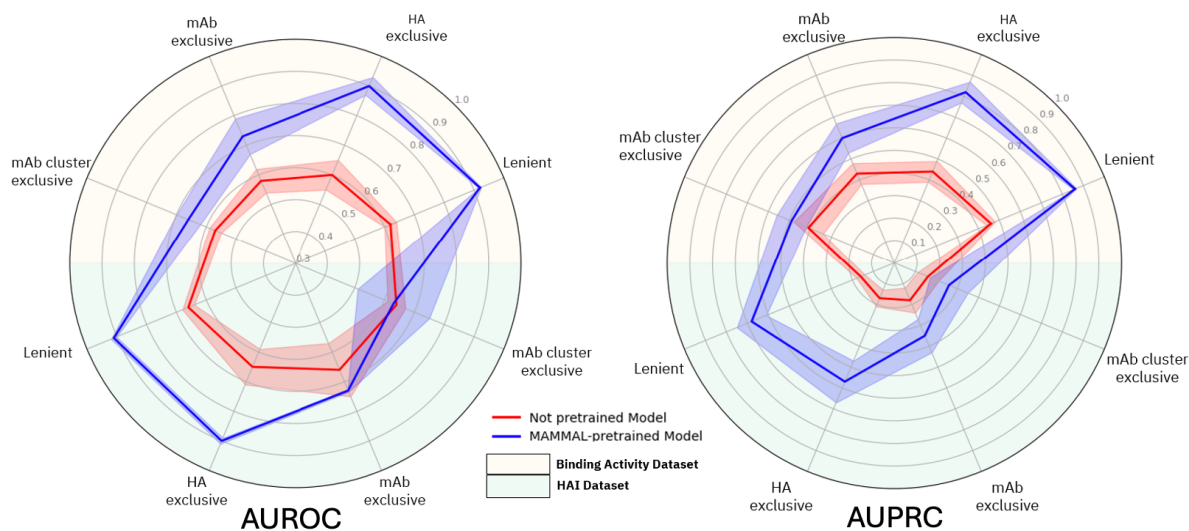

**Figure 2: Evaluation of binding activity and hemagglutination inhibition (HAI) classification models under various experimental conditions using the AUROC and AUPRC metrics**. Performance of fine-tuned MAMMAL models (blue) is compared to randomly initialized models (red). Solid lines represent the mean, while shaded areas indicate the standard deviation across the five folds in 5-fold cross-validation.

## Prediction of binding activity and hemagglutination inhibition

We conducted 8 experiments using 5-fold cross-validation, corresponding to two classification outcomes—binding and HAI—and four data splits (lenient, HA-exclusive, mAb-exclusive, and mAb-cluster-exclusive). Each experiment was repeated twice: once with random weight initialization (the "random-initialization" model) and once with initialization using MAMMAL weights (the "MAMMAL-finetuned" model). The evaluation metrics for these experiments are presented in Figure 2 and Table 4 in the appendix. Both the random-initialization and MAMMAL-finetuned models exhibited performance metrics significantly higher than random, where a random classifier is expected to produce an AUROC of 0.5 for both datasets and AUPRCs of 0.35 and 0.11 for the binding and HAI datasets, respectively.

Comparing the two models, the MAMMAL-finetuned model consistently outperformed the random-initialization model, except for the AUROC metric on the mAb-cluster-exclusive split for the HAI classification task. However, this difference fell within the range of the estimated standard errors (SEs, calculated as the standard deviation divided by the square root of 5) and was therefore not statistically significant (P = 0.43, one-sided t-test). Given that the HAI dataset is imbalanced (with only an 11% positivity rate), the AUPRC metric may be more relevant [86]. A significant difference was observed in AUPRC, with the MAMMAL-finetuned model showing superior performance.

For the MAMMAL-finetuned model in isolation, performance was higher in the lenient and HA-exclusive splits but lower in the mAb-exclusive and mAb-cluster-exclusive splits. These findings suggest that the model generalizes more effectively to unseen HA sequences than to unseen mAb sequences.

We assessed the robustness of the model's AUROC measures across various HA subgroups. As shown in **Error! Reference source not found.**3, the trend of higher performance in the lenient and HA-exclusive splits, coupled with reduced performance in the mAb-exclusive splits, was consistent across all subgroups. Performance within each subgroup remained largely stable, with the exception of poorer results on (1) binding prediction for mouse antibodies and (2) HAI prediction for HAs from earlier years in the mAb-exclusive splits.

## Prediction of antibody breadth of protection

We assessed the models' ability to predict antibody breadth for binding activity and HAI across the HA subtypes H1 and H3. The analysis was conducted separately for each HA subtype (H1 and H3) and assay type (binding activity and HAI). Antibodies featured in fewer than five assays were filtered out. For each antibody, we measured its breadth of protection by aggregating assay results and calculating the proportion of positive assays. We then computed a prediction score for broad protection by averaging the predictions from validation folds across all assays in which the antibody appears. Only antibody-exclusive splits were considered, as the analysis aimed to predict the broad protection of unseen antibodies.

To evaluate the predictive power of the aggregated model scores, we calculated the Pearson correlation between these scores and the antibodies' rates of positive assay results. Additionally, we calculated the AUROC for scores associated with at least 30% positive assays. Table 4 presents antibody statistics for each assay type (binding activity or HAI) and HA subtype (H1 and H3), along with the Pearson correlation and AUROC metrics.

**Table 3: Analysis of antibody breadth of protection: statistics and prediction metrics**

| task | HA subtype | mAb split | #mAbs | pearson | p-value | positive assays rate > 0.3 | AUROC |
|---|---|---|---|---|---|---|---|
| Binding Activity prediction | H1 | exclusive | 145 | 0.45 | 2.E-08 | 59% | 0.73 |
| | | cluster exclusive | 145 | 0.25 | 0.003 | 64% | 0.69 |
| | H3 | exclusive | 101 | 0.32 | 0.001 | 54% | 0.68 |
| | | cluster exclusive | 101 | 0.25 | 0.01 | 40% | 0.62 |
| HAI prediction | H1 | exclusive | 142 | 0.26 | 0.001 | 5% | 0.73 |
| | | cluster exclusive | 142 | 0.24 | 0.005 | 5% | 0.64 |
| | H3 | exclusive | 110 | 0.49 | 6.E-08 | 28% | 0.72 |
| | | cluster exclusive | 110 | 0.14 | 0.1 | 34% | 0.53 |

(a)

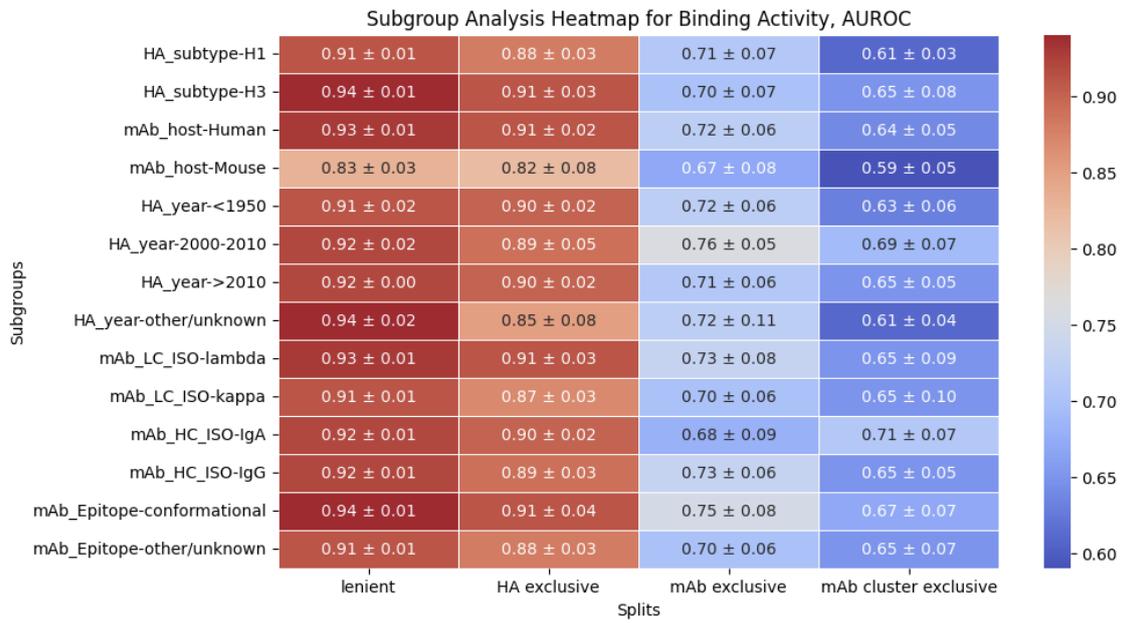

(b)

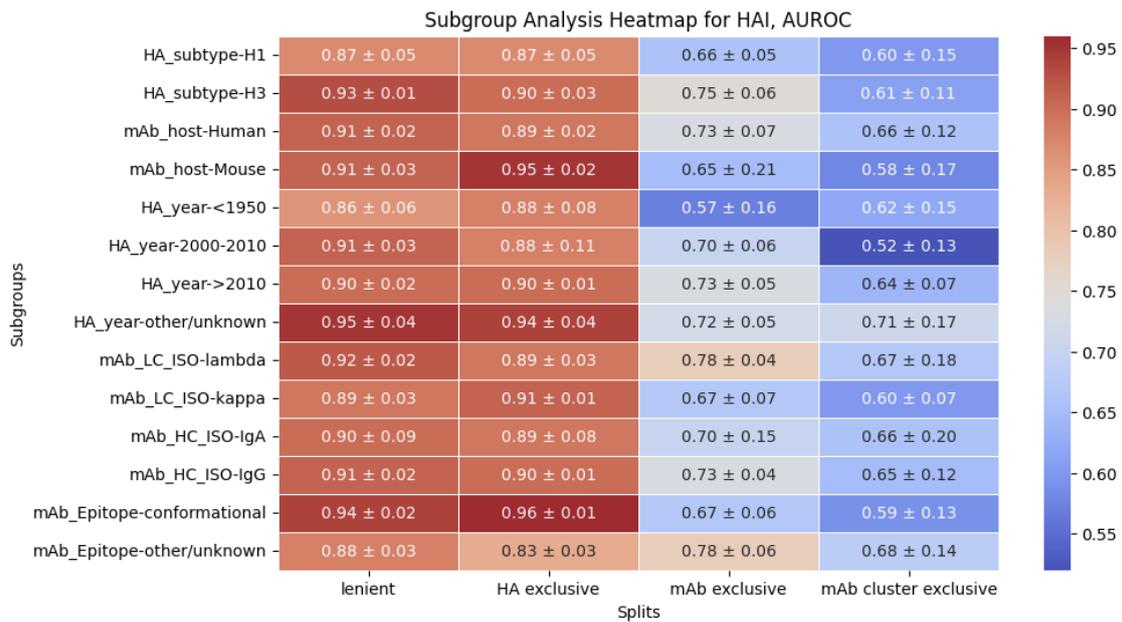

**Figure 3: Subgroup analysis for binding activity (a) and HAI (b).** The AUROC metric was calculated as an average over 5 folds of predictions for each of the subgroups. This metric is applied to each of the four splits evaluated in this study. Subgroup names combine a data characteristic name (e.g., *HA_subtype*) with its category value (e.g., *H1*).

## Discussion

In this study, we developed and evaluated AI models to predict antibody binding and HAI activity on influenza A HA using antibody HC and LC variable region sequences as input. Our approach leveraged MAMMAL, a language model pre-trained on extensive biomedical datasets, which we fine-tuned using laboratory-derived antibody-HA binding and HAI assay data. We evaluated model performance through comprehensive cross-validation analyses, employing multiple split paradigms to simulate diverse real-world scenarios and assess the robustness of our models.

To address the computational challenges posed by high-dimensional sequence data, we employed transformer architectures [87], which excel at capturing complex patterns in protein sequences data [28,27,88]. The superior performance of models initialized with MAMMAL weights compared to those with random initialization underscores the value of transfer learning from large-scale pre-trained biomedical models, particularly when working with limited task-specific data. These results underscore the transformative potential of pretraining on large-scale protein and antibody datasets to boost model performance on small, specialized datasets.

The high predictive performance (AUROC > 0.9) of our fine-tuned MAMMAL models under lenient split conditions highlights those models' ability to generalize across randomly held-out antibody-HA pairs. This result implies potential utility in reducing experimental workloads by predicting outcomes for untested combinations within known sequence spaces. Similarly, the strong performance (AUROC = 0.9) in the HA-exclusive split scenario demonstrates reliable prediction capabilities for novel HA sequences against previously analyzed antibodies, offering support for strain surveillance and antibody evaluation against emerging variants.

The more challenging antibody-exclusive split scenario, relevant for novel antibody design, showed moderate performance (AUROC = 0.73) for both binding activity and HAI prediction. This decrease in performance compared to other scenarios suggests current limitations in generalizing predictions to entirely new antibody sequences. The further reduction in performance under the HA-cluster-exclusive split (AUROC = 0.66 for binding, 0.63 for HAI) highlights particular challenges in extrapolating to more divergent HA sequences. Despite these limitations, the models showed promising results toward identifying broadly protective antibodies, especially against H3 subtypes (AUROC 0.64-0.73). Performance limitations likely arise because of the scarcity of characterized mAbs in datasets used in this study.

A key factor to improving model performance regards the expansion of training datasets through automated collection and curation of influenza neutralization assays from public repositories. Training on larger datasets with more diverse antibody sequence data is likely needed to enhance model generalization, particularly for novel antibody sequences. Future work will focus on the inclusion of a higher number of mAb sequences and datasets, including those from publicly available antibody databases (e.g., SAbDab [89], PLAbDab [90], IEDB [91]). However, the inclusion of datapoints from other groups may present challenges tied to lack of harmonization in methodologies used to determine the biological activities of mAbs (e.g., binding and HAI activity) and inconsistencies in units of measurement across datasets. Laboratory validation of in silico predictions is essential to advance AI models' performance on novel antibody-HA pairs. Such validation can also feedback to informing the training process and could contribute to an iterative approach to creating a more robust and accurate predictive frameworks.

# Conclusion

Our findings demonstrate the potential of fine-tuned language models for predicting antibody-HA interactions across various practical scenarios. While performance on novel antibodies, particularly those divergent from training data, remains an area for improvement, these models hold significant promise for accelerating influenza research and antibody design. Integration with computational antibody design pipelines, including AI-based systems, could enable rapid *in-silico* assessment of candidate antibodies and significantly expedite the antibody design process.

# Acknowledgement

This work was supported by the National Institute of Allergy and Infectious Diseases (NIAID), a component of the U.S. National Institutes of Health (NIH), Department of Health and Human Services, under contract 75N93019C00052 and by the Cleveland Clinic Foundation.

# Appendix

**Table 4: Evaluation of binding and HAI classification models.** For each experiment (task + data split) and metric, we compare random weight initialization ("random-initialization") with MAMMAL weight initialization ("MAMMAL-finetuned"). Bolded values indicate the higher performance for each comparison.

| Task | Data Split | Random-Initialization Model | | MAMMAL-finetuned Model | |
|---|---|---|---|---|---|
| | | AUROC | AUPRC | AUROC | AUPRC |
| Binding Prediction | Lenient | 0.62 ± 0.02 | 0.46 ± 0.02 | **0.92** ± 0.004 | **0.86** ± 0.007 |
| | HA exclusive | 0.60 ± 0.05 | 0.44 ± 0.05 | **0.90** ± 0.03 | **0.82** ± 0.05 |
| | mAb exclusive | 0.58 ± 0.04 | 0.43 ± 0.05 | **0.73** ± 0.06 | **0.60** ± 0.07 |
| | mAb cluster exclusive | 0.57 ± 0.02 | 0.41 ± 0.07 | **0.66** ± 0.04 | **0.49** ± 0.07 |
| HAI Prediction | Lenient | 0.66 ± 0.02 | 0.16 ± 0.02 | **0.91** ± 0.009 | **0.68** ± 0.07 |
| | HA exclusive | 0.65 ± 0.06 | 0.17 ± 0.04 | **0.90** ± 0.01 | **0.57** ± 0.10 |
| | mAb exclusive | 0.66 ± 0.09 | 0.18 ± 0.06 | **0.73** ± 0.04 | **0.35** ± 0.08 |
| | mAb cluster exclusive | **0.64** ± 0.03 | 0.16 ± 0.05 | 0.63 ± 0.12 | **0.26** ± 0.09 |